\documentclass[conf]{new-aiaa} 
\usepackage[utf8]{inputenc}
\usepackage{textcomp}
\usepackage{multicol}
\usepackage{lscape}
\usepackage[capitalise]{cleveref}
\usepackage{algorithm}
\usepackage{algpseudocode}
\usepackage{graphicx}
\usepackage{amsmath}
\usepackage{siunitx}
\usepackage{longtable,tabularx}
\usepackage{xcolor} 
\usepackage{wrapfig} 
\usepackage{subcaption} 
\usepackage{array}
\newcolumntype{L}[1]{>{\raggedright\let\newline\\\arraybackslash\hspace{0pt}}m{#1}}
\newcolumntype{C}[1]{>{\centering\let\newline\\\arraybackslash\hspace{0pt}}m{#1}}
\newcolumntype{R}[1]{>{\raggedleft\let\newline\\\arraybackslash\hspace{0pt}}m{#1}}

\usepackage[frozencache,cachedir=.]{minted}
\title{CoRL: Environment Creation and Management Focused on System Integration\footnote{Approved for Public Release. Case Numbers: APRS-RYZ-2023-01-00006}}
\author{Justin D. Merrick\footnote{Deputy Branch Chief, Autonomous Controls Branch, 21300 8th St.} and Benjamin K. Heiner\footnote{Behavior Development Lead AACO, Autonomy Capability Team 3, 2241 Avionics Circle.}}
\affil{Air Force Research Laboratory, Wright-Patterson AFB, OH, 45433}
\author{Cameron Long\footnote{Analyst, 6800 Cortona Dr}, Brian Stieber\footnote{Deputy Director, Algorithm/AI Solutions, 6800 Cortona Dr} and Steve Fierro\footnote{Senior Analyst, 6800 Cortona Dr}}
\affil{Toyon Research Corporation, Goleta, CA, 93117}
\author{Vardaan Gangal\footnote{Junior Autonomy Engineer,1415 Research Park Dr}}
\affil{Jacobs, Dayton, OH, 45432}
\author{Madison Blake\footnote{Senior Staff Engineer, 209 Madison St.}}
\affil{Heron Systems, Alexandria, VA, 22314}
\author{Joshua Blackburn\footnote{Lead Engineer, 2611 Commons Blvd, Ste 150}}
\affil{STR, Beavercreek, OH, 45431}

\begin{document}
\maketitle

\begin{abstract}
Existing reinforcement learning environment libraries use monolithic environment classes, provide shallow methods for altering agent observation and action spaces, and/or are tied to a specific simulation environment. The Core Reinforcement Learning library (CoRL) is a modular, composable, and hyper-configurable environment creation tool. It allows minute control over agent observations, rewards, and done conditions through the use of easy-to-read configuration files, pydantic \cite{pydantic} validators, and a functor design pattern. Using integration pathways allows agents to be quickly implemented in new simulation environments, encourages rapid exploration, and enables transition of knowledge from low-fidelity to high-fidelity simulations. Natively multi-agent design and integration with Ray/RLLib \cite{pmlr-v80-liang18b} at release allow for easy scalability of agent complexity and computing power. The code is publicly released and available at \url{https://github.com/act3-ace/CoRL}.
\end{abstract}

\section{Introduction}
Existing Reinforcement Learning (RL) software libraries can be roughly broken into three categories and their combinations: environment API, algorithm implementation, and integration. Integration handles the communication between the previous two. The introduction of the OpenAI Gym environment API \cite{Brockman2016} led to a proliferation of software packages utilizing this API \cite{gym-dependencies}. Despite this popularity, the Gym API has significant limitations. Most importantly, it lacks effective tooling for environment creation and alteration and does not natively support environments with multiple agents, which has led to the creation of multi-agent specific environment APIs such as PettingZoo \cite{terry2020pettingzoo}. The Core Reinforcement Learning library (CoRL) straddles the environment API and integration categories, providing a modular, composable, and hyper-configurable suite of tools for environment creation\footnote{Code available at: \url{https://github.com/act3-ace/CoRL}\label{foot:repo}}. CoRL allows users to rapidly explore complex design spaces with minimal re-integration. The contributions of CoRL are as follows:
\begin{enumerate}
    \item A suite of environment and agent creation tools,
    \item A hyper-configurable environment enabling rapid exploration for task-based RL,
    \item A flexible framework for developing these environments and the agents that populate them,
    \item Tooling to validate configuration files and give useful feedback when files are mis-configured, and
    \item Integration pathways enabling a dramatic reduction in development time to transition agents into a new simulation.
\end{enumerate}

The remainder of this paper compares CoRL to existing RL libraries (\cref{sec:rel_works}), expands upon the design of CoRL and how it achieves its environment creation capabilities (\cref{sec:design}), presents case studies highlighting the composable and modular nature of CoRL (\cref{sec:case_study}), and discusses its potential impacts and future improvements (\cref{sec:conclusions}).

\section{Related Works}\label{sec:rel_works}
Gym \cite{Brockman2016} is perhaps the best known environment API library currently in use. Its quantity and variety of environments and its common sense API have made it ubiquitous. This has also resulted in a large number of RL algorithms and RL algorithm libraries being designed for use with Gym. Stable Baselines \cite{stable-baselines3}, Tianshou \cite{tianshou}, and RLLib \cite{pmlr-v80-liang18b} are just a couple of the many libraries which have implementations that are utilize this API. Gym organizes each environment into a single class structure, implementing four main methods and six attributes. Of principle importance are the \texttt{step} method, which takes an action as an argument and runs one time step of the environment's dynamics, returning the next observation, the reward for the time step, whether the episode is complete, and an information dictionary that varies per environment; the \texttt{reset} method, which sets the environment to an initial state and returns an initial observation and an information dictionary; the \texttt{action\_space} attribute which describes the type, cardinality, and bounds of the actions required by the environment; and the \texttt{observation\_space} attribute, which returns the type, cardinality, and bounds of the observations output by the environment. CoRL uses this interface in its own Environment class and includes full integration with existing environments that use the Gym API. However, CoRL eschews the monolithic class structure for a more modular framework that allows users to easily modify their environments via YAML configuration files. The wrappers used to modify existing Gym environments have a similar function to these configuration management tools in CoRL. However, the modular and configurable nature of observations, rewards, and dones in CoRL allow complex permutations of a single task, and implementations of new tasks, in a way that is quick, readable, and repeatable.

Petting Zoo \cite{terry2020pettingzoo} is another environment API which extends the capabilities of Gym to multiple agents and introduces the Agent Environment Cycle Game in which agents act and receive reward in sequential manner. CoRL is also natively multi-agent and capable of handling agent death and exposing complex reward attribution. Unified Distributed Environment (UDE) \cite{la2022unified} is also multi-agent capable and seeks to improve on the modularity of simulation environments through its bridge component. This is similar in concept to the CoRL Simulator class, but UDE does not seek to create a configurable and composable environment and instead is merely providing an integration tool. UDE also has environment virtualization capabilities, similar to RLLib, which CoRL does not seek to emulate. Gymnasium \cite{gymnasium} is a fork of Gym with a slightly different API that differentiates between "terminated", meaning an agent met its goal or violated a constraint, and "truncated", meaning an episode met its time limit without reaching its goal or violating a constraint. These semantics are mirrored in CoRL through the use of a \texttt{DoneStatusCode} and a dictionary of Done functors.

The Unity ML-Agents Toolkit \cite{Unity} spans all three categories of library, but in a more closed ecosystem. It uses a similar structure to and implements many of the features of CoRL. The Agents in the ML-Agents Learning Environment are analogous to CoRL Platforms, while the Behaviors are analogous to instances of the Policy class. Sensors exist in both CoRL and Unity ML-Agetns in a very similar implementation. This level of modularity is also extended to the done state and the rewards, allowing access to the rewards and done states of individual agents. The differences between Unity ML-Agents and CoRL lie in the ability of CoRL to use any simulation environment, at the cost of some integration work, and in the Glue structure of CoRL being much more flexible and composable than the Sensor structure in Unity ML-Agents. CoRL also does not attempt to implement any RL algorithms, relying on other libraries to implement those in a manner consistent with the Environment class being used.

The DeepMind Control Suite (\texttt{dm\_control}) \cite{tunyasuvunakool2020dm_control} is a popular environment creation tool that pairs the MuJoCo \cite{todorov2012mujoco} physics library with a Python front-end that allows user to assemble new environments and tasks in ways very similar to CoRL. The XML specifications of the environment used in \texttt{dm\_control} provide precise control over the physics and objects in an environment, while CoRL's YAML configuration files provide the same precise control and composability over the observations, rewards, actions, and dones. \texttt{dm\_control} provides similar precise control over rewards as well by the implementation of their \texttt{tolerance()} function. CoRL takes a different approach in implementing some simple reward classes and providing a framework for the user to implement their own reward functions. The API utilized by \texttt{dm\_control} is different from that of Gym, which CoRL utilizes. The return from the \texttt{step} function returns information about the type of episode termination, similar to that of Gymnasium, which CoRL implements with its \texttt{DoneStatusCode}. The focus of \texttt{dm\_control} is on the creation of a simulation, while the focus of CoRL is on the management and manipulation of the outputs of that simulation. The slightly different API used by \texttt{dm\_control} means that fewer algorithm libraries support its use or do so through the use of wrappers which enforce conformity with the Gym API.

SuperSuit \cite{SuperSuit} is an integration tool between environments and algorithms that implements a number of more complex wrappers to Gym environments, such as frame stacking or delaying observations. All of these wrappers use the Gym or Gymnasium API and their functions can be or are replicated in CoRL.

\section{Design}\label{sec:design}
\begin{figure}
    \centering
    \includegraphics[width = 1.0\textwidth]{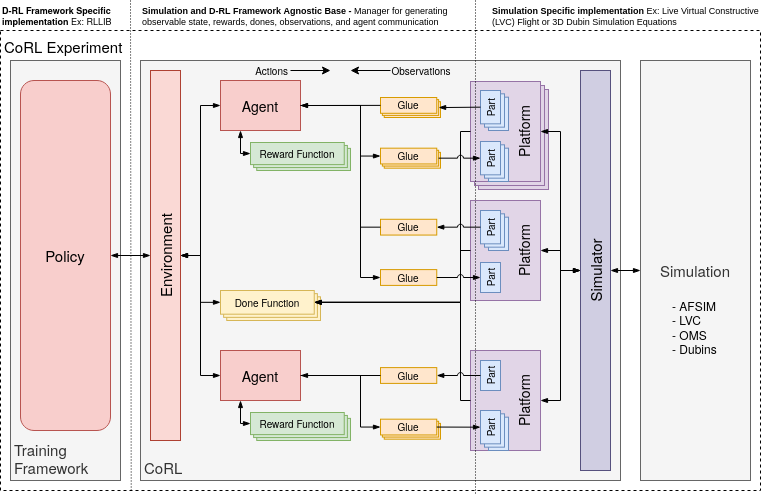} 
    \caption{A high Level view of CoRL showing the default information flows. (from left to right) The Training Framework is external to CoRL and must be compatible with the Environment class. The Environment class houses and manages all other classes. The Agent instances house Reward Functors that utilize the observations provided by Glues to construct the observation passed to the Training Framework. Glues utilize the measurements of the Sensors of Platforms or pass the controls to the Platform's Controller(s). Done Functors evaluate Platform state and report the done status to the Environment. Platforms are managed by the Simulator, which passes data to and from the Simulation, which is external to CoRL.}
    \label{fig:corl-high-level-structure}
\end{figure}

CoRL is designed to be modular, composable, and hyper-configurable. It accomplishes these goals through the use of pydantic\cite{pydantic} validators which specify the construction of YAML configuration files, a stateful functor design pattern for Glues, Rewards, and Termination/Goal Critiera (i.e. Dones), an Agent and Platform abstraction which allow hierarchies of agents to emerge naturally, and a pluggable simulator interface. An environment class is used to manage each of these constructs and ensure final compatibility with the training algorithm's interface. \cref{fig:corl-high-level-structure} shows a high level view of CoRL.

\subsection{Validators and Functors}\label{sec:val_func}
Each class in CoRL has its own validator. These serve three purposes. The first is to provide typing requirements for the class's attributes. These serve to guide a user in constructing a YAML configuration file for the class. Examples of this pairing between validator and configuration file are shown in \cref{sec:case_study}. Second, the validator is used to perform functional checks on the inputs provided. Third, the validators are used to interpret the configuration arguments to generate values and complex datatypes that enable the composability of CoRL Glues, Rewards, and Dones. This last purpose is accomplished through the use of the Episode Parameter Provider (EPP), the Reference Store (\cref{sec:epp_ref}), and CoRL's unit library. The EPP and Reference store supply values from a distribution to the functors. The unit library automatically converts units based on the configuration arguments. Specifically, the units of a functor must match the type (e.g. velocity, distance) of unit used in any wrapped or referenced classes. In addition to implementing a number of base and derived physical units, CoRL also implements and enforces fractions and percentages, as well as a \texttt{None} unit type.

The stateful functor design pattern used for CoRL's Glues, Rewards, and Dones combines naturally with the validators to allow a single configuration file entry to direct the composition of multiple functors. These create a directed acyclic graph of functors, which is used to reduce the number of instances of identical functors. Additionally, this functor design allows the user to operate at their preferred level of detail. A single functor can be created to do a complex manipulation of platform state, or a chain of functors can be used to accomplish the same calculation, allowing reuse. The statefulness of CoRL functors also allows communication to flow between the three main types of functors (Glues, Rewards, and Dones). Each of these classes and their interactions is explained in greater detail in \cref{sec:glues,sec:rew_done}.

\subsection{Simulator}\label{sec:simulator}
The Simulator class allows the CoRL Environment class (\cref{sec:environment}) to modularly interact with a variety of simulations of varying complexity with minimal reintegration. At its core a Simulator class looks very similar to a Gym Environment, implementing a \texttt{reset} and \texttt{step} method to reset or advance the simulation dynamics; however, the Simulator has attributes for \texttt{frame\_rate}, \texttt{sim\_time}, and \texttt{platforms}. These attributes allow for the creation, destruction, or unresponsiveness of Platforms (\cref{sec:platforms}) to be communicated to the Environment. The similarity to the Gym API allows for environment dynamics to be implemented directly in the Simulator class for simple problems or where no external simulation is present or necessary.

The minimal reintegration is achieved through a plugin library of Platform Parts. This allows for a rapid and principled transition from virtual simulation, to bench testing, to real-world testing. A Simulator class is defined for a virtual simulation, with a separate Simulator defined for both bench testing and real-world testing. By implementing the same types of Parts for each Simulator, the engineer can be confident that the measurements that the Glues create are consistent across these implementations. 

\subsection{Platforms}\label{sec:platforms}
Platforms are an important piece of the modularity of CoRL. A platform is an abstraction of a piece of the environment that has a measurable or modifiable state. In the simplest implementations, this can be the entire environment. CoRL uses this implementation for default interaction with existing Gym environments. A more physical interpretation might be in an orbital dynamics problem where a platform is defined as a spacecraft. In this example, the Platform class can be used to define the core spacecraft states such as its position and velocity, as well as how to extract them from the Simulator class. 

\begin{figure}
    \centering
    \includegraphics[width=\textwidth]{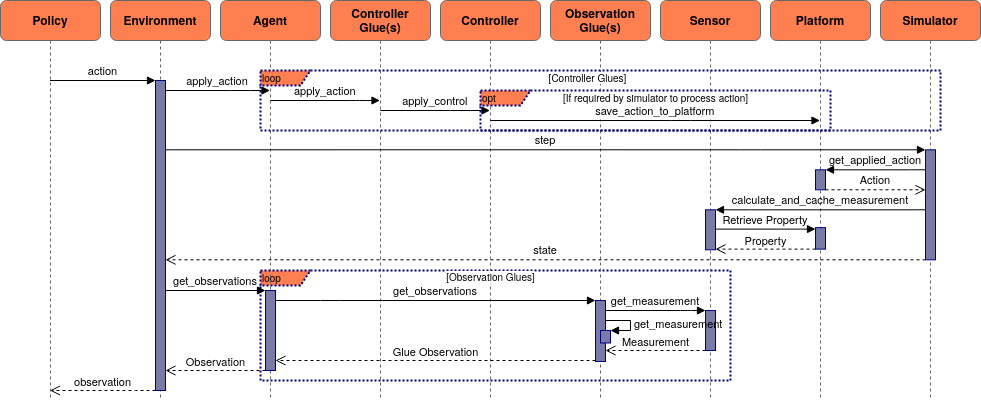}
    \caption{Sequence diagram for creation of an observation in the \texttt{step} function of CoRL's environment}
    \label{fig:obs_sequence}
\end{figure}

In all cases, Platforms have some combination of Sensors and Controllers, which are collectively called Parts. Parts dictate how the observation or modification of Platform states are achieved, with Sensors observing the state of the platform for downstream use by Glues and Controllers allow modifying the controllable states of the Platform in order to affect its future state, as determined by the dynamics of the Simulator. \cref{fig:obs_sequence} shows how this sequence is used in the \texttt{step} method of the environment to construct the observation for a single agent occupying a single platform.

In order to minimize re-integration efforts in the case where the simulation dynamics change or the platform changes, Parts have properties which define the limits, cardinality, and units of the space in which they are valid, similar to the spaces provided in Gym. Additionally, Parts are added to a plug-in library. This library registers a Part with a specific group name and conditions that must be met to invoke the Part. For example, a condition might require a specific Platform type or Simulator class be used. The functionality behind these constructs is that Parts can be added to an Agent's configuration file using the group name. That group name is then referenced with the plug-in library and the conditions for using the Parts in that group are checked until a match is found. Then, if the Simulator or Platform is changed, the Agent configuration file need not change, as the plug-in library search will select the appropriate Part to attach to the Platform or Platforms the Agent is attached to. This modularity of Parts also creates a natural pathway to dealing with asynchronous or missing measurement problems in a localized fashion. CoRL Sensors implement a value hold by default in the case of missing or malformed measurements.

\subsection{Episode Parameter Provider and Reference Store}\label{sec:epp_ref}
The Episode Parameter Provider (EPP) enables the environment to modify its parameters, such as the initial state, on a per-episode basis throughout the training process. These modifications can happen using two separate mechanisms. First, the parameters returned by the EPP are actually distributions rather than values, which the environment then samples to get the parameter value.  This allows the user to interject randomness into the training process by using different values each episode. CoRL natively provides parameters that implement constant, uniform, truncated Gaussian, and discrete choice distributions. Furthermore, there is an abstract interface that allows users to implement other distributions by extending the base parameter class.

Second, the EPP contains hooks that allow it to modify the distribution of parameters as training progresses. When defining the original parameter distributions, the user can specify updaters on the distribution hyperparameters, for example to increment a hyperparameter by a fixed step size. At the end of each training iteration, the environment passes the training result to the episode parameter provider. This allows it to modify its internal data, such as calling the updater on the parameter distributions that it will return to the environment. By creating subclass implementations of the episode parameter provider, the user is able to implement various forms of curriculum learning, such as domain randomization \cite{tobin2017, peng2018} or automatic domain randomization \cite{openai_adr}.  

As an example use of the EPP, consider an extension to Cartpole \cite{Barto1983NeuronlikeAE} where the height of the pole could be varied for each episode. In order to solve Cartpole with an arbitrary height, the training might use the episode parameter provider to specify this height as a uniform parameter with some bounds. Furthermore, the upper bound on the distribution could grow larger using an updater as part of a curriculum learning scheme.

The EPP is able to provide parameters to glues, rewards, dones, simulators, and simulator platform initialization. If multiple elements require the same parameter (such as a done termination value and a reward that computes distance from that termination value), the parameter can be put into the Reference Store. Then the relevant objects can all reference the same value after the parameter has been sampled by the environment. The EPP is implemented as a Ray Actor \cite{ray2.0, moritz2018ray}, meaning CoRL has a strong dependency on Ray.

\subsection{Glues}\label{sec:glues}
Glues are a flexible integration class used to communicate and transform information from the Platforms (and their Parts) to the Agents (\cref{sec:agents-policys}) and vice versa. When communicating from Platform to Agent, the glue instances construct the observation that the Agent will use to calculate the rewards and done, and which the Policy will use to determine what action(s) to take. When communicating from Agent to Platform, the glue instances construct the action that the Platform, and ultimately the Simulator, can use to propagate the dynamics.

Due to their composable nature, Glues can vary greatly in complexity and may use the environment state, platform state and sensors, reference store parameters, and other glues in their calculations. Several base types of Glues are included in CoRL. For the most basic functionality, the \texttt{ObserveSensor} and \texttt{ControllerGlue} Glues connect Platform Sensors and Controllers to the Agent and the Policy. To aid in Glue composition, wrapper Glues that can wrap one or more other Glues and, in the case of wrapping multiple Glues, store them in list or dictionaries, are also included. Other basic mathematical functions such as unit vector transformation, norm calculation, projection, and differencing are included. Examples Glues and their compositions are shown in \cref{sec:case_study}.

To take advantage of the directed acyclic graph created by wrapped Glues (and eventually by Rewards and Dones, \cref{sec:rew_done}, as well), Glues may optionally use an Extractor class. The extractor simplifies that configuration YAML and is used to topologically sort the dependencies of the CoRL functors. This reduces memory requirements and improves computation speed by eliminating identical instances of a Glue.

\subsection{Rewards and Dones}\label{sec:rew_done}

\begin{figure}
    \centering
    \includegraphics[width = 1.0\textwidth]{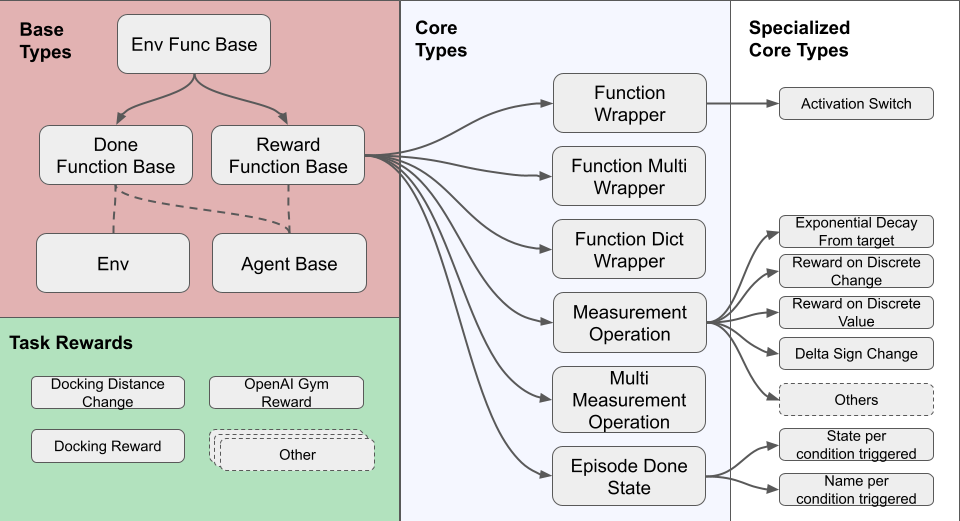} 
    \caption{CoRL Reward types: (1) Base types - general API shared by all rewards, (2) Task Rewards - Specific rewards utilized by the environments in \cref{sec:case_study}, (3) Core Types - General task independent rewards, (4) Specialized Core Types - Specific reward implementations that provide shaping or advanced logic for activation.}
    \label{fig:functors}
\end{figure}

Dones are a class of functors that test whether their criteria have been met and, if so, mark their scope done and issue a status code to indicate whether this constitutes a win, partial win, draw, partial loss, or loss. The scope of a particular Done can be a single agent or the environment as a whole. These different scopes allow for a variety of done conditions to hold. For instance, the default behavior in CoRL does not end an episode until each agent has had a Done evaluate True. Through a configuration file, this can be changed to end an episode as soon as any agent is done. Furthermore, with environment Dones, called Shared Dones and which have a slightly different call signature, the episode can be terminated regardless of the done status of particular agents. 

Similar to Dones, Rewards are a class of functors that, when evaluated, produce a scalar value for a particular component of an agent's reward. The rewards for each agent are eventually summed in the Environment to produce the scalar reward that the Gym API expects, but, by calculating the component rewards individually, and reporting them to Tensorboard, analysis of an agent's general behaviors can be conducted analyzed across training, similar to the process proposed in \cite{macglashan2022value}. Rewards are specifically calculated after Dones in order to allow rewards to utilize the done status code created by any Done functor that evaluates to True.

\begin{figure}
    \centering
    \includegraphics[width=\textwidth]{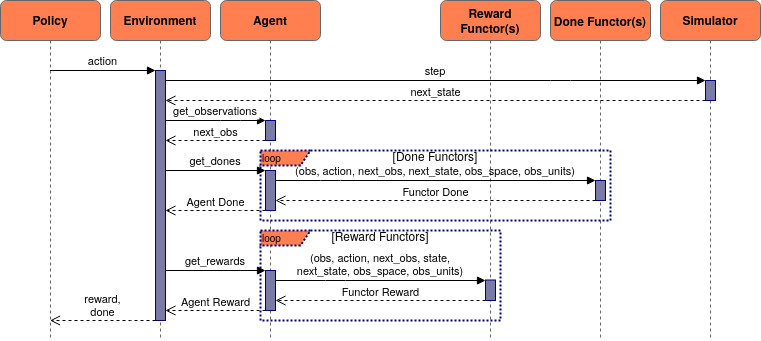}
    \caption{Sequence diagram for creation of rewards and dones in the \texttt{step} function of CoRL's environment}
    \label{fig:rew_done_sequence}
\end{figure}

As with all functors in CoRL, both Rewards and Dones are composable and configurable. \cref{fig:functors} shows some examples of Rewards and Dones and how they can compose. \cref{fig:rew_done_sequence} shows a sequence diagram of the processing of the Dones and Rewards as implemented in CoRL's Environment class (\cref{sec:environment}). Similar to Glues, Rewards and Dones may also take advantage of the graph of functors through the Extractor class.

\subsection{Agents and Policies}\label{sec:agents-policys}
The CoRL Agent class implements an agent as described in \cite{russell2010artificial}. Each instance of an agent may partially or fully control any number of platforms and contains a dictionary of Glues, which serve as the agent's sensors and actuators. The instance also contains a dictionary of Rewards, which the agent uses to communicate with a Policy defined in its configuration. The Policy serves to map the Glue observations to an action. 

The structure of a Policy class must be dictated by the algorithm implementation library and the environment API that CoRL is integrating with. While integration with RLLib is included in the CoRL release, a Policy implementation is not limited to that library, nor must it be compatible with RL in general. Examples of a random action policy and a scripted policy are included in the CoRL release. 

This construction of an agent as a separate entity from a Platform or a Policy allows a natural progression from single-agent to multi-agent environments, both competitive and cooperative. It also allows for shared Policys among Agents, multiple Agents utilizing the same Platform, and a robust means of developing hierarchies of Agents. Furthermore, it minimizes the re-integration necessary for transferring agents from learning environments to inference environments, or from simple dynamics to complex dynamics.

\subsection{Environment}\label{sec:environment}
At the opposite end of CoRL from the Simulator is the Environment class, which integrates the Agents, Platforms, Simulator, EPP, Reference Store, Glues, Dones, and Rewards into a cohesive whole that complies with the interfaces expected by the training algorithm. Included in the CoRL release is a multi-agent environment that complies with the the RLLib MultiAgentEnv API, a multi-agent version of the Gym API. While the configuration files specify how the environment is to be created, the Environment class processes these configurations to initialize the component parts of the environment. 

The environment is also responsible for maintaining any global Parameters and Dones that are specified in the configuration, such as Parameters that are applicable to multiple agents, or Dones that deal with the overall state of the simulation as opposed to any particular Agent or Platform. A simple example of this is a hard environment horizon. If the episode is terminated after a fixed number of steps, the meta nature of this done condition means it is best placed in the environment. In addition, the Environment also checks the observation(s) obtained from the Agent(s) to ensure that the measurements lie within the space defined by the Glues. This sanity check can be configured to check every step, or it can be used to spot-check observations throughout training.  

\subsection{Visualization/Debugging Tools} 
CoRL, through its Ray/RLLib integration, utilizes Tensorboard \cite{tensorflow2015-whitepaper} to log and track metrics and generate visualizations for them. Within CoRL, Tensorboard has been configured to log and track all rewards, dones, and reference store parameters. Additionally, the \texttt{DoneStatusCode} and simulator reset parameters are also logged. All of these measurements are automatically aggregated to maximum, minimum and mean. All parameters are also logged to a CSV file for ease of visualization outside of Tensorboard.

The configuration of RLLib, the Environment, the Agents, and the Policies are also saved for each trial in order to assist in debugging or determining the differences between runs. These are stored in a JSON file. As some parameters may change over the course of training, especially those modified by an EPP, the information logged to Tensorboard, along with the configuration of the agent and environment for each iteration are logged in a JSON file.

\subsection{Evaluation Framework}\label{sec:evaluation}
After a policy has been trained, the Evaluation Framework can be used to investigate the policy's performance. The evaluation framework follows the same CoRL design practices, utilizing YAML configuration files and a functor design pattern. The framework has been constructed to enable three core processes: Evaluate, Generate Metrics, and Visualize. Each process utilizes separate YAML configuration files to supply the appropriate inputs and each may be run separately or in series as an end-to-end pipeline.

\begin{figure}
    \centering
    \includegraphics[width=1.0\textwidth]{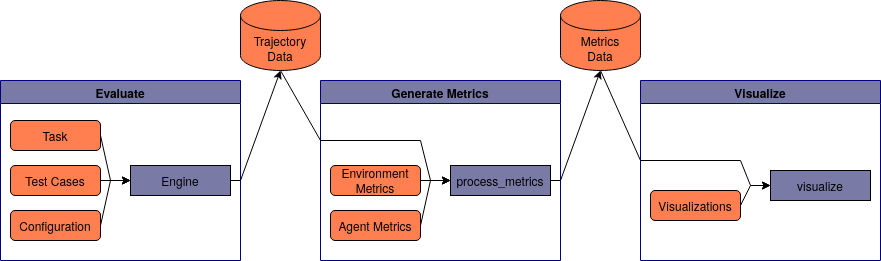}
    \caption{The Evaluation Framework}
    \label{fig:evaluate}
\end{figure}

The Evaluate process executes rollouts from a user-defined set of initial conditions. To achieve this, the evaluation framework captures CoRL agent and environment definitions and uses them with an engine to execute rollouts. The engine is specific to the training framework used. At release, CoRL implements an RLLib engine that is compatible with all Policy classes included in CoRL. The trajectory resulting from a rollout is contained in an \texttt{EpisodeArtifact} class which stores agent information such as observations, rewards, dones, and platform state, as well as environment state information. An \texttt{EpisodeArtifact} is generated and saved to disk for each member of the set of initial conditions.

The Generate Metrics process computes metrics from the trajectory data collected and stored to disk by the Evaluate process. Metric classes using the same composable functor design pattern as Glues, Rewards, and Dones define the quantities to be computed. These quantities can be computed directly from a trajectory, can be aggregated across all trajectories, or can be aggregated over quantities generated by other metrics. Some example metrics are the number of times an agent successfully solves the task and the proportion of total reward each reward functor contributed aggregated across a trajectory. The evaluation framework provides a layer of abstraction for the type of quantity returned from a metric functor, categorizing them into two types, terminal and non-terminal. A terminal metric can be categorized as one which can be represented as a non-container type (e.g. float, string, integer). A non-terminal metric returns a container type (e.g. dictionary, list). This distinction is made to assist in compatibility with non-Python visualizations. A handful of common abstract metrics are implemented at release, but a user can craft more specific metrics to suit their needs.

The Visualize process creates visualizations from the computed metrics. A generic interface is provided that can be implemented to provide a custom visualization. There are two visualizations provided, a visualization that prints the metrics computed in a table to stdout and another that creates HTML plots of the metrics. Multiple visualizations can be included in a single configuration file for the Visualize process.

\section{Case Study}\label{sec:case_study}
Two case studies are included to illustrate the design principles of CoRL. The first uses the OpenAI Gym CartPole environment as a simulation and uses CoRL's \texttt{GymSimulator} and its associated classes to demonstrate the modularity and composability of CoRL Glues, Rewards, and Dones. The second is an abstraction of the spacecraft docking problem in which the degrees of freedom have been limited to one dimension for simplicity. This case study shows how a simulation environment can be integrated into CoRL without first being set up as a Gym environment.

\subsection{CartPole and OpenAI Gym}\label{sec:cartpole}
Several variations on an agent designed to solve the CartPole-v1 Gym environment are included in CoRL to provide concrete examples of the Gym integration and Glue composability discussed in \cref{sec:design}. The simplest example is shown in \cref{lst:cartpole}. Here, a trainable agent (Line 1) is configured with two parts, one sensor and one controller (Lines 3-6). The \texttt{Sensor\_State} notation in the configuration file will trigger a search of the plug-in library for a Sensor with that name, which will yield a sensor that reports the Gym environment state as output by the \texttt{step} function. A baseline EPP is included which does not have any parameters (Line 7). Glues are added for each of the parts (Lines 8-24), each specifying which part they are connecting to. The \texttt{ObserveSensor} Glue is also configured to not normalize the observations pulled from the \texttt{Sensor\_State} sensor. Finally, Dones and Rewards are added from other files. Both \texttt{baseline\_dones.yml} and \texttt{baseline\_rewards.yml} contain a simple done and reward that report the done and reward provided by the Gym environment.

\begin{listing}
\begin{minted}{yaml}
"agent": "corl.agents.base_agent.TrainableBaseAgent"
"config": {
    "parts": [
        {"part": "Controller_Gym"},
        {"part": "Sensor_State"},
    ],
    "episode_parameter_provider": !include baseline_epp.yml,
    "glues": [
        {
            "functor": "corl.glues.common.controller_glue.ControllerGlue",
            "config": {
                "controller": "Controller_Gym",
            },
        },
        {
            "functor": "corl.glues.common.observe_sensor.ObserveSensor",
            "config":{
                "sensor": "Sensor_State",
                "normalization": {
                  "enabled": False
                }
            }
        },
    ],
    "dones": !include baseline_dones.yml,
    "rewards": !include baseline_rewards.yml,
}
\end{minted}
\caption{CoRL agent file for CartPole-v1 Gym environment}
\label{lst:cartpole}
\end{listing}

From this basic design it is simple to add complexity to the observation. For instance, \cref{lst:target_diff} shows the YAML configuration for adding a Glue which takes the zero-th index of the state and subtracts it from a target value, in this case zero, to return the difference between that state and the target value. Lines 5-8 are functionally identical to Lines 16-20 of \cref{lst:cartpole}, but in this case, the \texttt{TargetValueDifference} Glue has wrapped the \texttt{ObserveSensor} Glue in a dictionary with the key "sensor" (Line 3-4). This configuration also includes information on the resulting observation's units ("N/A") and the minimum and maximum measurements expected and allowed, which are used to construct the Glue's observation space (Lines 16-20).

\begin{listing}
\begin{minted}{yaml}
{
    "functor": "corl.glues.common.target_value_difference.TargetValueDifference",
    "wrapped": {
        "sensor": {
            "functor": "corl.glues.common.observe_sensor.ObserveSensor",
            "config":{
                "sensor": "Sensor_State",
                "normalization": {"enabled": False}
            }
        },
    },
    "config":{
        "target_value": 0,
        "index": 0,
        "unit": N/A,
        "limit": {
            "minimum": -5000,
            "maximum": 5000,
            "unit": N/A
        }
    },
}
\end{minted}
\caption{Example of Glue composability using the \texttt{TargetValueDifference} and \texttt{ObserveSensor} Glues}
\label{lst:target_diff}
\end{listing}

Complexity can also be added to the rewards. \cref{lst:exp_decay_rew} shows the configuration for a Reward which uses the Glue from \cref{lst:target_diff} and grants an exponentially decaying reward for how far the observation of the Glue is from a target value. This example makes use of the observation extractors discussed in \cref{sec:rew_done} to specify which values, spaces, and units to use (Lines 5-7). In this case, the Glue from \cref{lst:target_diff} is used. While that observation is already a difference to a target value, the modularity of the Rewards and Glues allow the same measurement to be constructed entirely in the reward, or for a different target value than zero to be rewarded. In this case, the default for the \texttt{target\_value} field is zero and it is not included. Lines 9-12 specify shaping parameters for the exponential decay as well as how to handle situations where the observation at time $t+1$ is farther away than the observation at time $t$. 

\begin{listing}
\begin{minted}{yaml}
{
    "name": "OpenAIGymExtractorReward",
    "functor": "corl.rewards.exponential_decay_from_target_value.ExponentialDecayFromTargetValue",
    "config": {
        "observation": {
            "fields": ["ObserveSensor_Sensor_StateDiff", "direct_observation_diff"]
        },
        "index": 0,
        "eps": 1, # THIS IS RADIANS
        "reward_scale": .000000000001,
        "is_closer": true,
        "closer_tolerance": 10,
    }
}
\end{minted}
\caption{Example of Reward complexity using the \texttt{ExponentialDecayFromTargetValue} Reward}
\label{lst:exp_decay_rew}
\end{listing}

Other variations on the CartPole agent in \cref{lst:cartpole} are included in the CoRL repository which highlight additional composability in the Glues and Rewards. Additionally, examples of non-operable agents, repeated observations, and agents without the ability to affect the environment are demonstrated. The integration of Gym wrappers and the use of keyword arguments passed to Gym environments is also demonstrated.

\subsection{1D Docking}\label{sec:docking}
The 1D Docking problem is implemented with the same structure as used in \cite{sa-dynamics,saferl-benchmarks} but represented as a simple linear ODE, 

\begin{equation}
    \dot{\mathbf{x}} = A\mathbf{x} + B\mathbf{u}
    \label{eq:linear_ode}
\end{equation}

where the state $\mathbf{x} = \left[x, \dot{x}\right]$, the position and velocity of the docking craft, the control $\mathbf{u} = [T]$, the thrust of the docking craft, and 
\begin{eqnarray}
    A = \begin{bmatrix}
        0 & 1 \\
        0 & 0
    \end{bmatrix} & 
    B = \begin{bmatrix}
        0 \\
        \frac{1}{m}
    \end{bmatrix},
\end{eqnarray}
the state and input matrices.

This implementation uses the concept of an Entity class (\texttt{Deputy1d}) which is roughly analogous to a Platform in CoRL and a Dynamics class (\texttt{Docking1dDynamics}) which calculates $\dot{\mathbf{x}}$. The task of implementing this simulation in CoRL then becomes the creation of a Simulator, a Platform, and several Parts. Implementing \texttt{Docking1dSimulator} requires creating concrete methods for several abstract methods defined in the CoRL \texttt{BaseSimulator}, most importantly a \texttt{reset} and a \texttt{step} method. \cref{lst:docking_sim_reset} shows the \texttt{reset} method. Some of this method, particularly Lines 2-3 and 17-19, will be typical of most CoRL Simulator classes. Lines 6-10, however, interface directly with the simulation by constructing and storing the Entity class(es) associated with the simulation while Lines 12-15 pair the Entity class(es) with CoRL Platform(s), in particular the \texttt{Docking1dPlatform} created for this environment.

\begin{listing}
\begin{minted}{python}
def reset(self, config):
    config = self.get_reset_validator(**config)
    self.clock = 0.0

    # construct entities ("Gets the platform object associated with each simulation entity.")
    self.sim_entities = {}  # pylint: disable=attribute-defined-outside-init
    for agent_id, agent_config in self.config.agent_configs.items():
        agent_reset_config = config.platforms.get(agent_id, {})
        self.sim_entities[agent_id] = Deputy1D(name=agent_id, **agent_reset_config)

    # construct platforms ("Gets the correct backend simulation entity for each agent.")
    sim_platforms = {}
    for agent_id, entity in self.sim_entities.items():
        agent_config = self.config.agent_configs[agent_id]
        sim_platforms[agent_id] = Docking1dPlatform(platform_name=agent_id, platform=entity, parts_list=agent_config.parts_list)

    self._state = BaseSimulatorState(sim_platforms=sim_platforms, sim_time=self.clock)
    self.update_sensor_measurements()
    return self._state
\end{minted}
\caption{\texttt{reset} method of 1D Docking Simulator}
\label{lst:docking_sim_reset}
\end{listing}

The \texttt{Docking1DPlatform} requires a \texttt{Deputy1D} Entity in its validator, and uses the properties of this Entity as its own properties to pass along the state of the environment. \texttt{Docking1DPlatform} also stores the actions given to it as shown in \cref{fig:obs_sequence}. This allows the \texttt{step} function of the Simulator class, shown in \cref{lst:docking_sim_step} to retrieve these applied actions (Line 3), apply them to the appropriate Entity classes (Lines 4-5) and then update the sensor measurements on each Platform with the updated state of the Entity (Line 7).

\begin{listing}
\begin{minted}{python}
def step(self):
    for agent_id, platform in self._state.sim_platforms.items():
        action = np.array(platform.get_applied_action(), dtype=np.float32)
        entity = self.sim_entities[agent_id]
        entity.step(action=action, step_size=self.config.step_size)
        platform.sim_time = self.clock
    self.update_sensor_measurements()
    self.clock += self.config.step_size
    self._state.sim_time = self.clock
    return self._state
\end{minted}
\caption{\texttt{step} method of 1D Docking Simulator}
\label{lst:docking_sim_step}
\end{listing}

Sensor measurements are calculated from Sensor classes which access the sensor and velocity properties of their parent Platform. These sensors are added to the plug-in library and associated with \texttt{Docking1dSimulator} and \texttt{Docking1dPlatform}. Finally, a Controller is also created which wraps the \texttt{Docking1dPlatform} methods to save and get actions to the abstract Controller class methods. This Part is similarly added to the plug-in library.

Now a working interface between the simulation and the CoRL Environment and Agent classes has been made. It can be configured with YAML files similar to those in \cref{sec:cartpole}. A baseline configuration is included in the CoRL release. More detail on the Simulator, Platform, Part interfaces can be found in the source code or documentation (\cref{foot:repo}).

\section{Conclusions}\label{sec:conclusions}

CoRL is a modular, composable, and hyper-configurable environment creation tool. While some existing tools rely on monolithic implementations, CoRL allows for minute control of an agent's observations and actions with modification of YAML configuration files. Other tools are tied to specific simulation environments, CoRL allows for new simulations to be integrated with minimal difficulty. Still others are tied only to a reinforcement learning paradigm, CoRL can be used with RL agents trained using any framework, but can also utilize the distributed computing of Ray to evaluate non-RL agents with ease.

At release, CoRL provides useful examples and common implementations of many of its classes, but can be easily extended to more advanced features. \cref{sec:epp_ref} discussed using the EPP to do curriculum learning, but that framework could be expanded further to include automatic domain randomization. The EPP could also be upgraded to allow the reference store to provide information to any object in the environment, currently it only allows for functors to reference it. The existing unit conversion system is very rigid. Using an existing unit conversion library could allow users to define their own units. RLLib \cite{pmlr-v80-liang18b} is incompatible with the DeepMind Control Suite \cite{tunyasuvunakool2020dm_control}, but implementing a Simulator, Platform, and basic Glues could allow its popular baseline tasks to be manipulated by CoRL. CoRL is being actively maintained and used for a variety of research tasks at the Air Force Research Laboratory.

\bibliography{main}

\end{document}